\documentclass[english]{article}
\usepackage[T1]{fontenc}
\usepackage[latin9]{inputenc}
\usepackage[letterpaper]{geometry}
\usepackage{babel}
\usepackage{array}
\usepackage{verbatim}
\usepackage{float}
\usepackage{booktabs}
\usepackage{multirow}
\usepackage{amsmath}
\usepackage{amssymb}
\usepackage[unicode=true,pdfusetitle,
 bookmarks=true,bookmarksnumbered=false,bookmarksopen=false,
 breaklinks=false,pdfborder={0 0 1},backref=false,colorlinks=false]
 {hyperref}

\makeatletter

\providecommand{\tabularnewline}{\\}

\date{}
\usepackage{bbm}
\usepackage{graphicx}
\usepackage{setspace}

\makeatother

\begin{document}
\title{A Hidden Variables Approach to Multilabel Logistic Regression}
\author{Jaemoon Lee and Hoda Shajari\thanks{The authors are with the Department of Computer and Information Science
and Engineering, University of Florida, Gainesville, FL, USA. E-mail:
j.lee1,shajaris@ufl.edu.}}
\maketitle
\begin{abstract}
\noindent Multilabel classification is an important problem in a wide
range of domains such as text categorization and music annotation.
In this paper, we present a probabilistic model, Multilabel Logistic
Regression with Hidden variables (MLRH), which extends the standard
logistic regression by introducing hidden variables. Hidden variables
make it possible to go beyond the conventional multiclass logistic
regression by relaxing the one-hot-encoding constraint. We define
a new joint distribution of labels and hidden variables which enables
us to obtain one classifier for multilabel classification. Our experimental
studies on a set of benchmark datasets demonstrate that the probabilistic
model can achieve competitive performance compared with other multilabel
learning algorithms.
\end{abstract}

\section{Introduction\label{sec:1}}

Classification is one of the most widespread and classical supervised
learning problems in machine learning. Classification problems exist
quite extensively in many domains and research areas. In single-label
classification tasks, it is assumed that each instance belongs to
one and only one class and therefore is assigned a single label. The
goal is to learn a mapping from given data (training data) to their
respective labels which is able to predict the label of unseen instances
(test data). There have been a number of approaches to tackle multiclass
classification tasks, including bayesian and probabilistic approaches,
support vector machines, and artificial neural networks.

Although multiclass classification has pervasive applications and
has been applied successfully to many real-world problems, there exist
learning tasks which do not fit in multiclass learning framework.
In constrast to multiclass (single-label) classification, in multilabel
setting an instance could belong to multiple classes simultaneously
and therefore, more than one label needs to be assigned to each instance.
For example, a news document could be categorized as both \emph{politics}
and \emph{economics} labels simultaneously. In fact, a multilabel
classification problem can be decomposed into a number of independent
binary classification problems. However, it might not be optimal to
solve binary classification problems independently since some labels
might be correlated to each other. Therefore, multilabel classification
is more challenging. Some of these challenges include exponential
growth of possible number of label combinations, label dependencies
and structured output spaces \cite{MaximumMargin-Lampert:2011}, unbalanced
datasets \cite{Addressing-Imbalance-charte2015}, and computational
cost of developing and training these models. Multilabel classification
is related to many real-world applications such as text categorization
\cite{Text-Joachims1998,Text-McCallum1999,Text-Schapire2000,Text-Kazawa:2004,Text-Rousu:2006},
image and scene classification \cite{Image-Boutell2004,Image-Wang2008},
and multimedia automatic annotation and suggestion systems.

Logistic regression (LR) is one of the most common and well-known
approaches to solve the multiclass classification tasks and has proven
its value in the statistics and machine learning community. Standard
LR assumes that only one elements of a label vector is $1$ and others
are $0$ (one-hot-encoding) since it uses a multinoulli distribution
\cite{Murphy:2012}. Thus, it naturally fits the multiclass setting.
We estimate the parameters of LR based on maximum likelihood estimate
and predict a label of an instance in a probabilistic way. However,
in multilabel setting, one-hot encoding is the core difficulty for
extending LR to a multilabel classification framework. The reason
is that we can not assign multiple $1$s to a label simultaneously
because the multinoulli distribution is no longer a valid probability
distribution. In spite of this limitation of LR, a number of approaches
using LR for multilabel classification has been proposed in the literature.
Those approaches used independent binary LRs and combined other methods
such as \emph{k}-nearest neighbor (KNN) to consider correlations among
labels. Despite all the existing approaches for multilabel logistic
regression, no framework has been proposed for training the classifiers
within one model.

\noindent In this paper, we propose Multilabel Logistic Regression
with Hidden variables (MLRH), which extends standard LR by introducing
hidden variables. A hidden variable or a latent variable is a variable
which is not directly observable and affects the response variable.
Hidden variables are sometimes used to explain observed variables
or relationships between variables. We propose a novel joint probability
model of hidden variables and labels to go beyond the conventional
classification tasks.

The rest of this paper is organized as follows: Related work is discussed
in Section \ref{sec:2}. Our MLRH method is then described in Section
\ref{sec:3}. We briefly discuss the optimization algorithm in Section
\ref{sec:4} and related experiments are described in Section \ref{sec:5}.
The paper ends with a conclusion and future work in Section 6. Some
of detailed derivations of our method are discussed in Appendices
\ref{sec:Appendices}.%

\section{Related Work\label{sec:2}}

As it was mentioned, there have been approaches in the past which
extend LR for multilabel classification. Cheng and H�llermeier \cite{MultilabelLogistic-Cheng2009}
proposed an approach to combine both logistic regression and instance-based
learning. Their IBLR model basically uses binary LRs for each class.
In addition, to consider correlations among labels, labels of neighboring
instances are weighted by KNN and combined with the parameter of each
LR. In other words, their model uses the labels of neighboring instances
as extra attributes in a logistic regression scheme and train one
classifier for each label. Bian et al. \cite{MultilabelLogistic-Bian2012}
presented the CorrLog model, which explicitly model pairwise correlation
between labels and combine it with independent LR for each class.
In the CorrLog model, pairwise correlation is defined as weighted
sum of products between labels and is added to the parameter of each
LR. Li et al. \cite{MultilabelLogistic-Bian2016} combined the CorrLog
model with elastic-net regularization. Liu et al. \cite{MultilabelLogistic-LIU2014}
proposed MLSLR model, in which independent LRs are combined with elastic-net
regularization for each class and then they are trained. However,
the MLSLR do not consider possible label dependencies among labels.
Teisseyre \cite{MultilabelLogistic-Teisseyre2017} propose similar
approaches to the MLSLR model, but Teisseyre considers label depedencies
by suggesting chaining rule of probabilities of each label. These
approaches and some other approaches \cite{MLLogistic-PatentTask,MLLogistic-mPLR}
are based on using separate LRs for each class.

Our approach is based on a hidden variable model explained in details
in the next section for extending multiclass logistic regression to
multilabel case. There also have been models based on hidden variables
in logistic regression which are mainly concerned with conventional
multiclass (single-label) classification problem. Memisevic et al.
\cite{Logistic-Gated2010} proposed the Gated Softmax model, which
adopt mixture model and predicts labels by summing over all possible
configurations of hidden variables. In the Gated Softmax model, hidden
variables are combined with each input feature quadratically. Xu et
al. \cite{HiddenLogistic-xu2014architectural,Logistic-Xu2016} proposed
Multinomial Latent Logistic Regression (MLLR) model. The MLLR model
also asscociates hidden variables with each input feature. These approaches
basically stem from the idea that the probability $\Pr\left(y\mid\boldsymbol{x},\boldsymbol{W}\right)$,
which we want to estimate, can be obtained from a joint probability
of hidden variables and labels by marginalizing over all possible
configurations of hidden variables, i.e.,

\begin{equation}
\Pr\left(y\mid\boldsymbol{x},\boldsymbol{W}\right)=\sum_{\boldsymbol{h}}\Pr\left(y,\boldsymbol{h}\mid\boldsymbol{x},\boldsymbol{W}\right).
\end{equation}

\noindent The Gated Softmax and MLLR model use the following conditional
probability,

\begin{equation}
\Pr\left(y,\boldsymbol{h}\mid\boldsymbol{x},\boldsymbol{W}\right)=\frac{\exp\left\{ f\left(y,\boldsymbol{h},\boldsymbol{x},\boldsymbol{W}\right)\right\} }{\sum_{y,\boldsymbol{h}}\exp\left\{ f\left(y,\boldsymbol{h},\boldsymbol{x},\boldsymbol{W}\right)\right\} }.\label{eq:2}
\end{equation}

Our model is also based on the conditional probability (Eq. \ref{eq:2}).
However, in contrast to the previous approaches, we will suggest a
new joint conditional probability that can be applied to both multiclass
and multilabel setting.

\section{Multilabel Logistic Regression with Hidden Variables\label{sec:3}}

In this section, we briefly review multiclass logistic regression
and then propose our joint probability model of hidden variables and
labels for multilabel logistic regression. We demonstrate that multiclass
logistic regression can be obtained from this joint probability model
and show that multilabel logistic regression can also be achieved
from the joint probability model.

\subsection{Multiclass Logistic Regression\label{subsec:3.1}}

In the multiclass classification problem with $K$ classes $\left(K\geq2\right)$,
we are given $N$ training data $\left\{ \left(\boldsymbol{x}_{i},\boldsymbol{y}_{i}\right)\right\} _{i=1}^{N}$
with $\boldsymbol{x}_{i}\in\mathbb{R}^{D}$ and $\boldsymbol{y}_{i}\in\left\{ 0,1\right\} ^{K}$,
where $\sum_{k=1}^{K}y_{ik}=1$ for each $i$ (one-hot-encoding).
The goal is to learn a model from the training data such that given
a new test input $\boldsymbol{x}_{t}$, it can predict the label vector
$\boldsymbol{y}_{t}$. Logistic regression is one of the well-known
approaches for this task. Logistic regression originates from the
idea of applying regression model to classification by setting the
log-odds to be linear function of parameters and features \cite{Murphy:2012},
i.e.,

\begin{equation}
\text{log-odds=}\text{log}\frac{\text{Pr}\left(y_{k'}=1\mid\boldsymbol{x},\boldsymbol{W}\right)}{\text{Pr}\left(y_{K}=1\mid\boldsymbol{x},\boldsymbol{W}\right)}=\boldsymbol{w}_{k'}^{T}\boldsymbol{x},\qquad k'\in\left\{ 1,2,\ldots K-1\right\} 
\end{equation}

\noindent where $\boldsymbol{w}$ is a weight vector or decision surface
between classes. Using the fact that sum of all the $K$ possibilities
is $1$ and if $\Pr\left(y_{K}=1\mid\boldsymbol{x},\boldsymbol{W}\right)$
is chosen to be the reference, we can obtain the probabilities for
each class,

\begin{equation}
\Pr\left(y_{k}=1\mid\boldsymbol{x},\boldsymbol{W}\right)=\frac{\exp\left(\boldsymbol{w}_{k}^{T}\boldsymbol{x}\right)}{\sum_{l}\exp\left(\boldsymbol{w}_{l}^{T}\boldsymbol{x}\right)},\qquad k\in\left\{ 1,2,\ldots K\right\} \label{eq:softmax}
\end{equation}

\noindent where $\boldsymbol{W}=\left(\boldsymbol{w}_{1},\boldsymbol{w}_{2},\ldots\boldsymbol{w}_{K}\right)\in\mathbb{R}^{D\times K}$
and $\boldsymbol{w}_{k}$ is the weight vector of \textit{k}-th class.
(Eq. \ref{eq:softmax}) is known as the \emph{softmax} function.

The linear function of parameters and features, $\boldsymbol{w}_{k}^{T}\boldsymbol{x}$,
can also view as a score function, or an activation function \cite{Bishop:2006},
i.e.,

\begin{equation}
s_{k}\left(\boldsymbol{x};\boldsymbol{W}\right)=\boldsymbol{w}_{k}^{T}\boldsymbol{x}.\label{eq:multiclass score func}
\end{equation}

\noindent We can also obtain the posterior probabilities $\Pr\left(\boldsymbol{y}\mid\boldsymbol{x}\right)$
(Eq. \ref{eq:softmax}) by exponentiating and normalizing (Eq. \ref{eq:multiclass score func}).
Note that a bias term \emph{b} can be added easily by augmenting an
extra dimension to $\boldsymbol{W}$ and $\boldsymbol{x}$. In the
multiclass logistic regression, the assumption is that every feature
vector $x$ belongs to one and only one class. Then, the probability
of the training set is

\begin{equation}
\Pr\left(\left\{ \boldsymbol{y}_{i}\right\} \mid\left\{ \boldsymbol{x}_{i}\right\} ,\boldsymbol{W}\right)=\prod_{i=1}^{N}\prod_{k=1}^{K}\mu_{ik}^{y_{ik}},\label{eq:multiclass training set prob}
\end{equation}

\noindent where $\mu_{ik}=\Pr\left(y_{ik}=1\mid\boldsymbol{x}_{i},\boldsymbol{W}\right)$.
Note that each instance $\boldsymbol{x}_{i}$ is drawn from an independent
multinomial distribution (but not identical). The optimal parameters
of multiclass logistic regression are obtained by minimizing the negative
log-likelihood,

\begin{equation}
\boldsymbol{W^{*}}=\arg\min_{\boldsymbol{w}}\left\{ -\log\Pr\left(\left\{ \boldsymbol{y}_{i}\right\} \mid\left\{ \boldsymbol{x}_{i}\right\} ,\boldsymbol{W}\right)\right\} .
\end{equation}

For a test instance $\boldsymbol{x}_{t}$, we assign the \emph{k}-th
label, $y_{k}$, to be $1$, if the posterior probability $\Pr\left(y_{tk}=1\mid\boldsymbol{x}_{t},\boldsymbol{W}\right)$
has the maximum value (winner-take-all).

\subsection{Hidden Variables}

Hidden variables, as described in Section \ref{sec:1}, can be used
to explain relationships between variables. A hidden variable model
is a model that learns a relationships between a set of observable
variables and a set of hidden variables. Many hidden variable models
have been proposed in the literature, and thus have a long history
\cite{Latent-LDA2003,Latent-Dumais2004,Latent-HCRF2007,Latent-Bartholomew2011}.
As we weill see, a hidden variable approach is a sufficient condition
to derive the standard logistic regression. Furthermore, hidden variables
enable us to go beyond multiclass setting by relaxing single label
constraint (one-hot-encoding) and accommodating logistic regression
to multi-label framework. We will derive the standard logistic regression
from first principles with hidden variables. Then, we discuss its
inherent problem and suggest our Multilabel Logistic Regression with
Hidden variables (MLRH) model. We introduce binary hidden variables
\textbf{$\boldsymbol{h}$},

\begin{equation}
\boldsymbol{h}=\left[h_{1},h_{2},\ldots,h_{K}\right],\qquad h_{k}\in\left\{ 0,1\right\} ,
\end{equation}

\noindent into the \emph{softmax} function (Eq. \ref{eq:softmax}).
Since we condiser the multiclass setting, we put the one-hot encoding
setting to the hidden variables \textbf{$\boldsymbol{h}$},

\begin{equation}
\sum_{k}h_{k}=1.\label{eq:const of hidden var}
\end{equation}

\noindent The joint probability model of hidden variables $\boldsymbol{h}$
and labels $\boldsymbol{y}$ is

\begin{equation}
\Pr\left(\boldsymbol{y},\boldsymbol{h}\mid\boldsymbol{x},\boldsymbol{W}\right)=\frac{1}{Z(\boldsymbol{x},\boldsymbol{W})}\exp\left[\sum_{k'}\left\{ y_{k'}\log h_{k'}+h_{k'}\boldsymbol{w}_{k'}^{T}\boldsymbol{x}\right\} \right],\label{eq:joint prob-mismatch}
\end{equation}

\noindent where

\begin{equation}
Z\left(\boldsymbol{x},\boldsymbol{W}\right)=\sum_{\boldsymbol{y},\boldsymbol{h}}\exp\left[\sum_{k'}\left\{ y_{k'}\log h_{k'}+h_{k'}\boldsymbol{w}_{k'}^{T}\boldsymbol{x}\right\} \right].\label{eq:norm constant-mismatch}
\end{equation}

This joint probability distribution is actually same with the joint
model (Eq. \ref{eq:2}) we discussed in Section \ref{sec:2}. By defining
$0\times\log0$ to be $1$, we can get back the standard logistic
regression model. We compute marginal probability $\Pr\left(\boldsymbol{y}\mid\boldsymbol{x},\boldsymbol{W}\right)$
over all the possible configurations of hidden variables \textbf{$\boldsymbol{h}$},

\begin{equation}
\Pr\left(\boldsymbol{y}\mid\boldsymbol{x},\boldsymbol{W}\right)=\sum_{\boldsymbol{h}}\Pr\left(\boldsymbol{y},\boldsymbol{h}\mid\boldsymbol{x},\boldsymbol{W}\right).
\end{equation}

\noindent For example, assume the number of classes is two ($K=2$).
After some algebra, we can easily obtain the follwing set of probabilities
which are identical to the softmax function (Eq. \ref{eq:softmax}),

\begin{equation}
\Pr\left(y_{1}=1,y_{2}=0\mid\boldsymbol{x},\boldsymbol{W}\right)=\frac{1}{Z\left(\boldsymbol{x},\boldsymbol{W}\right)}\exp\left(\boldsymbol{w}_{1}^{T}\boldsymbol{x}\right)=\frac{\exp\left(\boldsymbol{w}_{1}^{T}\boldsymbol{x}\right)}{\exp\left(\boldsymbol{w}_{1}^{T}\boldsymbol{x}\right)+\exp\left(\boldsymbol{w}_{2}^{T}\boldsymbol{x}\right)},
\end{equation}

\begin{equation}
\Pr\left(y_{1}=0,y_{2}=1\mid\boldsymbol{x},\boldsymbol{W}\right)=\frac{1}{Z\left(\boldsymbol{x},\boldsymbol{W}\right)}\exp\left(\boldsymbol{w}_{2}^{T}\boldsymbol{x}\right)=\frac{\exp\left(\boldsymbol{w}_{2}^{T}\boldsymbol{x}\right)}{\exp\left(\boldsymbol{w}_{1}^{T}\boldsymbol{x}\right)+\exp\left(\boldsymbol{w}_{2}^{T}\boldsymbol{x}\right)}.
\end{equation}

\noindent Therefore, we proved that the joint probability model (Eq.
\ref{eq:joint prob-mismatch}) of hidden variables and lables works
well in the multiclass setting.

However, there exists an inherent problem related to \emph{log} term
in the joint porbability, which we call as a \emph{mismatch} problem.
Consider the two-class case. The joint probability (Eq. \ref{eq:joint prob-mismatch})
can be simplified as

\begin{equation}
\Pr\left(\boldsymbol{y},\boldsymbol{h}\mid\boldsymbol{x},\boldsymbol{W}\right)=\frac{1}{Z(\boldsymbol{x},\boldsymbol{W})}h_{1}^{y_{1}}\exp\left(h_{1}\boldsymbol{w}_{1}^{T}\boldsymbol{x}\right)h_{2}^{y_{2}}\exp\left(h_{2}\boldsymbol{w}_{1}^{T}\boldsymbol{x}\right).
\end{equation}

\noindent In multiclass setting, $h_{1}^{y_{1}}\times h_{2}^{y_{2}}$
produce $1$ if both $h_{k}$ and $y_{k}$ have same value (e.g. $0^{0}\times1^{1}=1)$,
otherwise it becomes $0$. In other words, if $h_{k}$ and $y_{k}$
do not \emph{match}, the joint probability (Eq. \ref{eq:joint prob-mismatch})
becomes $0$. However, in the multilabel setting we should relax the
one-hot encoding setting of labels $\boldsymbol{y}$,

\begin{equation}
1\leq\sum_{k}y_{k}\leq K,
\end{equation}

\noindent where $K$ is the number of classes. To go beyond multiclass
setting, the constraint of hidden variables \ref{eq:const of hidden var}
is also relaxed,

\begin{equation}
\sum_{k}h_{k}\leq K.
\end{equation}

This relaxation causes the \emph{mismatch} problem to $h_{k}^{y_{k}}$.
Table \ref{tab:1} shows an example of the mismatch problem. Regardless
of values of labels $\boldsymbol{y}$, $h_{k}^{y_{k}}$ always produces
$1$ when every component of $\boldsymbol{h}$ is $1$. Thus, an extra
exponential term is always added in the joint probability (Eq. \ref{eq:joint prob-mismatch})
when every hidden variable is $1$, and thus hidden variables do not
reflect the information of labels correctly.

\noindent 
\begin{table}[H]
\caption{Mismatch problem in $h_{k}^{y_{k}}$\label{tab:1}}

\centering{}%
\begin{tabular}{cccccc}
\toprule 
{\small{}$y_{2}$} & {\small{}$y_{1}$} & {\small{}$h_{2}$} & {\small{}$h_{1}$} & {\small{}$h_{1}^{y_{1}}\times h_{2}^{y_{2}}$} & {\scriptsize{}$\exp\left[\sum_{k'}\left\{ y_{k'}\log h_{k'}+h_{k'}\boldsymbol{w}_{k'}^{T}\boldsymbol{x}\right\} \right]$}\tabularnewline
\midrule
\midrule 
\multirow{4}{*}{{\small{}0}} & \multirow{4}{*}{{\small{}1}} & {\small{}0} & {\small{}0} & {\small{}$0$} & {\small{}$0$}\tabularnewline
\cmidrule{3-6} \cmidrule{4-6} \cmidrule{5-6} \cmidrule{6-6} 
 &  & {\small{}0} & {\small{}1} & {\small{}$1$} & {\small{}$\exp\boldsymbol{w}_{1}^{T}\boldsymbol{x}$}\tabularnewline
\cmidrule{3-6} \cmidrule{4-6} \cmidrule{5-6} \cmidrule{6-6} 
 &  & {\small{}1} & {\small{}0} & {\small{}$0$} & {\small{}$0$}\tabularnewline
\cmidrule{3-6} \cmidrule{4-6} \cmidrule{5-6} \cmidrule{6-6} 
 &  & {\small{}1} & {\small{}1} & {\small{}$1$} & {\small{}$\exp\left(\boldsymbol{w}_{1}+\boldsymbol{w}_{2}\right)^{T}\boldsymbol{x}$}\tabularnewline
\midrule
\multirow{4}{*}{{\small{}1}} & \multirow{4}{*}{{\small{}0}} & {\small{}0} & {\small{}0} & {\small{}$0$} & {\small{}$0$}\tabularnewline
\cmidrule{3-6} \cmidrule{4-6} \cmidrule{5-6} \cmidrule{6-6} 
 &  & {\small{}0} & {\small{}1} & {\small{}$0$} & {\small{}$0$}\tabularnewline
\cmidrule{3-6} \cmidrule{4-6} \cmidrule{5-6} \cmidrule{6-6} 
 &  & {\small{}1} & {\small{}0} & {\small{}$1$} & {\small{}$\exp\boldsymbol{w}^{T}\boldsymbol{x}$}\tabularnewline
\cmidrule{3-6} \cmidrule{4-6} \cmidrule{5-6} \cmidrule{6-6} 
 &  & {\small{}1} & {\small{}1} & {\small{}$1$} & {\small{}$\exp\left(\boldsymbol{w}_{1}+\boldsymbol{w}_{2}\right)^{T}\boldsymbol{x}$}\tabularnewline
\bottomrule
\end{tabular}
\end{table}

\noindent Therefore, we modified the joint probability model (Eq.
\ref{eq:joint prob-mismatch}) to resolve the mismatch problem. Instead
of the exponential scheme of $h_{k}$ and $y_{k}$, we introduce the
multiplication scheme of $h_{k}$ and $y_{k}$, i.e.,

\begin{equation}
\Pr\left(\boldsymbol{y},\boldsymbol{h}\mid\boldsymbol{x},\boldsymbol{W}\right)=\frac{1}{Z\left(\boldsymbol{x},\boldsymbol{W}\right)}\prod_{k}\left\{ \frac{r_{k}+1}{2}\exp\left(h_{k}\boldsymbol{w}_{k}^{T}\boldsymbol{x}\right)\right\} ,\label{eq:new joint prob}
\end{equation}

\noindent where

\begin{equation}
r_{k}=\left(2h_{k}-1\right)\left(2y_{k}-1\right),\label{eq:multiplication scheme}
\end{equation}

\noindent and the normalization constant $Z\left(\cdot\right)$ is
computed as

\begin{equation}
Z\left(\boldsymbol{x},\boldsymbol{W}\right)=\sum_{\boldsymbol{y},\boldsymbol{h}}\left[\prod_{k}\left\{ \frac{r_{k}+1}{2}\exp\left(h_{k}\boldsymbol{w}_{k}^{T}\boldsymbol{x}\right)\right\} \right].\label{eq:new partition func}
\end{equation}

Note that hidden variables $\boldsymbol{h}$ and $\boldsymbol{y}$
are mapped to $\left\{ -1,1\right\} $ space in (Eq. \ref{eq:multiplication scheme}).
This prevents the result multiplication scheme to be $0$ because
even one $0$ can make entire result be $0$. Now their product $r_{k}$
produces $1$ If $h_{k}$ and $y_{k}$ are match, otherwise it becomes
$-1$, and therefore hidden variables reflect the information of labels
correctly. The multiplication scheme resolves the mismatch problem
and prevents the previous joint probability model (Eq. \ref{eq:joint prob-mismatch})
to produce biased results. In the modified joint probability model
(Eq. \ref{eq:new joint prob}), we remape the space of $r_{k}$ from
$\left\{ -1,1\right\} $ to $\left\{ 0,1\right\} $ to ensure that
all the probabilities are always greater than or equal to $0$. Since
the modified joint probability model is essentially same as the previous
model (Eq. \ref{eq:joint prob-mismatch}) in multiclass setting, we
can get back the standard logistic regression by marginalizing the
joint probability (Eq. \ref{eq:new joint prob}) over hidden variables.

\subsection{Multilabel Logistic Regression with Hidden Variables\label{subsec:3.3}}

In the multilabel classification with $K$ classes $\left(K\geq2\right)$,
we are given $N$ training data $\left\{ \left(\boldsymbol{x}_{i},\boldsymbol{y}_{i}\right)\right\} _{i=1}^{N}$
with $\boldsymbol{x}_{i}\in\mathbb{R}^{D}$ and $\boldsymbol{y}_{i}\in\left\{ 0,1\right\} ^{K}$.
But instead of $\sum_{k=1}^{K}y_{ik}=1$, $y_{ik}$ can have multiple
$1$s. Since logistic regression stems from multinoulli distribution,
multiclass logistic regression can not be naturally extended to solve
the multi-label problem by having multiple $1$s in label $\boldsymbol{y}$
because it will not be a valid probability distribution. However,
hidden variables enables to relax the one-hot-encoding setting and
thus allows us to go beyond multiclass problems. We first consider
a two-class multilabel classification problem, and then later consider
the general case. We enumerate all the possible configurations of
hidden variables $\boldsymbol{h}$ in the joint probability model
(Eq. \ref{eq:new joint prob}). Table \ref{tab:2} shows that from
equation \ref{eq:new partition func}, the partition function $Z\left(\boldsymbol{x},\boldsymbol{W}\right)$
becomes

\begin{equation}
Z\left(\boldsymbol{x},\boldsymbol{W}\right)=\exp\left(\boldsymbol{w}_{1}^{T}\boldsymbol{x}\right)+\exp\left(\boldsymbol{w}_{2}^{T}\boldsymbol{x}\right)+\exp\left\{ \left(\boldsymbol{w}_{1}+\boldsymbol{w}_{2}\right)^{T}\boldsymbol{x}\right\} .
\end{equation}

\noindent 
\begin{table}[H]
\noindent \begin{centering}
\begin{tabular}{>{\centering}p{3mm}>{\centering}p{3mm}>{\centering}p{3mm}>{\centering}p{3mm}c}
\toprule 
{\scriptsize{}$y_{2}$} & {\scriptsize{}$y_{1}$} & {\scriptsize{}$h_{2}$} & {\scriptsize{}$h_{1}$} & {\scriptsize{}$s(\boldsymbol{h},\boldsymbol{x},\boldsymbol{W})$}\tabularnewline
\midrule
\midrule 
\multirow{4}{3mm}{{\scriptsize{}0}} & \multirow{4}{3mm}{{\scriptsize{}1}} & {\scriptsize{}0} & {\scriptsize{}0} & {\scriptsize{}$0$}\tabularnewline
\cmidrule{3-5} \cmidrule{4-5} \cmidrule{5-5} 
 &  & {\scriptsize{}0} & {\scriptsize{}1} & {\scriptsize{}$\exp\left(\boldsymbol{w}_{1}^{T}\boldsymbol{x}\right)$}\tabularnewline
\cmidrule{3-5} \cmidrule{4-5} \cmidrule{5-5} 
 &  & {\scriptsize{}1} & {\scriptsize{}0} & {\scriptsize{}$0$}\tabularnewline
\cmidrule{3-5} \cmidrule{4-5} \cmidrule{5-5} 
 &  & {\scriptsize{}1} & {\scriptsize{}1} & {\scriptsize{}$0$}\tabularnewline
\bottomrule
\end{tabular}\quad{}%
\begin{tabular}{>{\centering}p{3mm}>{\centering}p{3mm}>{\centering}p{3mm}>{\centering}p{3mm}c}
\toprule 
{\scriptsize{}$y_{2}$} & {\scriptsize{}$y_{1}$} & {\scriptsize{}$h_{2}$} & {\scriptsize{}$h_{1}$} & {\scriptsize{}$s(\boldsymbol{h},\boldsymbol{x},\boldsymbol{W})$}\tabularnewline
\midrule
\midrule 
\multirow{4}{3mm}{{\scriptsize{}1}} & \multirow{4}{3mm}{{\scriptsize{}0}} & {\scriptsize{}0} & {\scriptsize{}0} & {\scriptsize{}$0$}\tabularnewline
\cmidrule{3-5} \cmidrule{4-5} \cmidrule{5-5} 
 &  & {\scriptsize{}0} & {\scriptsize{}1} & {\scriptsize{}$0$}\tabularnewline
\cmidrule{3-5} \cmidrule{4-5} \cmidrule{5-5} 
 &  & {\scriptsize{}1} & {\scriptsize{}0} & {\scriptsize{}$\exp\left(\boldsymbol{w}_{2}^{T}\boldsymbol{x}\right)$}\tabularnewline
\cmidrule{3-5} \cmidrule{4-5} \cmidrule{5-5} 
 &  & {\scriptsize{}1} & {\scriptsize{}1} & {\scriptsize{}$0$}\tabularnewline
\bottomrule
\end{tabular}\quad{}%
\begin{tabular}{>{\centering}p{3mm}>{\centering}p{3mm}>{\centering}p{3mm}>{\centering}p{3mm}c}
\toprule 
{\scriptsize{}$y_{2}$} & {\scriptsize{}$y_{1}$} & {\scriptsize{}$h_{2}$} & {\scriptsize{}$h_{1}$} & {\scriptsize{}$s(\boldsymbol{h},\boldsymbol{x},\boldsymbol{W})$}\tabularnewline
\midrule
\midrule 
\multirow{4}{3mm}{{\scriptsize{}1}} & \multirow{4}{3mm}{{\scriptsize{}1}} & {\scriptsize{}0} & {\scriptsize{}0} & {\scriptsize{}$0$}\tabularnewline
\cmidrule{3-5} \cmidrule{4-5} \cmidrule{5-5} 
 &  & {\scriptsize{}0} & {\scriptsize{}1} & {\scriptsize{}$0$}\tabularnewline
\cmidrule{3-5} \cmidrule{4-5} \cmidrule{5-5} 
 &  & {\scriptsize{}1} & {\scriptsize{}0} & {\scriptsize{}$0$}\tabularnewline
\cmidrule{3-5} \cmidrule{4-5} \cmidrule{5-5} 
 &  & {\scriptsize{}1} & {\scriptsize{}1} & {\scriptsize{}$\exp\left(\boldsymbol{w}_{1}+\boldsymbol{w}_{2}\right)^{T}\boldsymbol{x}$}\tabularnewline
\bottomrule
\end{tabular}
\par\end{centering}
{\footnotesize{}\caption{Summing over all configuration set of $\boldsymbol{h}$. Note that
$s\left(h,x,W\right)$ is $\prod_{k}\left\{ \frac{r_{k}+1}{2}\exp\left(h_{k}\boldsymbol{w}_{k}^{T}\boldsymbol{x}\right)\right\} $\label{tab:2}}
}{\footnotesize\par}
\end{table}

\noindent Then the marginal probabilities $\Pr\left(\boldsymbol{y}\mid\boldsymbol{x},\boldsymbol{W}\right)$
are

\begin{equation}
\Pr\left(y_{1}=1,y_{2}=0\mid\boldsymbol{x},\boldsymbol{W}\right)=\frac{1}{Z\left(\boldsymbol{x},\boldsymbol{W}\right)}\exp\left(\boldsymbol{w}_{1}^{T}\boldsymbol{x}\right),\label{eq:22}
\end{equation}

\begin{equation}
\Pr\left(y_{1}=0,y_{2}=1\mid\boldsymbol{x},\boldsymbol{W}\right)=\frac{1}{Z\left(\boldsymbol{x},\boldsymbol{W}\right)}\exp\left(\boldsymbol{w}_{2}^{T}\boldsymbol{x}\right),\label{eq:23}
\end{equation}

\begin{equation}
\Pr\left(y_{1}=1,y_{2}=1\mid\boldsymbol{x},\boldsymbol{W}\right)=\frac{1}{Z\left(\boldsymbol{x},\boldsymbol{W}\right)}\exp\left\{ \left(\boldsymbol{w}_{1}+\boldsymbol{w}_{2}\right)^{T}\boldsymbol{x}\right\} .\label{eq:24}
\end{equation}

\noindent Note that the joint probability model $\Pr\left(\boldsymbol{y},\boldsymbol{h}\mid\boldsymbol{x},\boldsymbol{W}\right)$
is valid in the multilabel setting since

\begin{equation}
\sum_{\boldsymbol{y}}\sum_{\boldsymbol{h}}\Pr\left(\boldsymbol{y},\boldsymbol{h}\mid\boldsymbol{x}_{i},\boldsymbol{W}\right)=\sum_{\boldsymbol{h}}\sum_{\boldsymbol{y}}\Pr\left(\boldsymbol{y},\boldsymbol{h}\mid\boldsymbol{x}_{i},\boldsymbol{W}\right)=1.
\end{equation}

\noindent The observation is that if $y_{k}=1$, then $\exp\left(\boldsymbol{w}_{k}^{T}\boldsymbol{x}\right)$
term is included in the marginal probability. We can obtain a new
probability distribution over label $\boldsymbol{y}$ with hidden
variables

\begin{align}
\Pr\left(\boldsymbol{y}\mid\boldsymbol{x},\boldsymbol{W}\right) & =\frac{1}{Z\left(\boldsymbol{x},\boldsymbol{W}\right)}\exp\left(y_{1}\boldsymbol{w}_{1}^{T}\boldsymbol{x}\right)\exp\left(y_{2}\boldsymbol{w}_{2}^{T}\boldsymbol{x}\right).
\end{align}

\noindent Then, the probability of the training set becomes

\begin{equation}
\Pr\left(\left\{ \boldsymbol{y}_{i}\right\} \mid\left\{ \boldsymbol{x}_{i}\right\} ,\boldsymbol{W}\right)=\prod_{i}\left\{ \frac{1}{Z\left(\boldsymbol{x}_{i},\boldsymbol{W}\right)}\exp\left(y_{i1}\boldsymbol{w}_{1}^{T}\boldsymbol{x}_{i}\right)\exp\left(y_{i2}\boldsymbol{w}_{2}^{T}\boldsymbol{x}_{i}\right)\right\} .\label{eq:27}
\end{equation}

\noindent Our objective function is the negative log-likelihood of
(Eq. \ref{eq:27}),

\begin{align}
\ell\left(\boldsymbol{W}\right) & =\sum_{i}-\log\Pr\left(\left\{ \boldsymbol{y}_{i}\right\} \mid\left\{ \boldsymbol{x}_{i}\right\} ,\boldsymbol{W}\right)\nonumber \\
 & =\sum_{i}\left[\log Z\left(\boldsymbol{x}_{i},\boldsymbol{W}\right)-y_{i1}\boldsymbol{w}_{1}^{T}\boldsymbol{x}_{i}-y_{i2}\boldsymbol{w}_{2}^{T}\boldsymbol{x}_{i}\right].\label{eq:28}
\end{align}

\noindent Note that the objective function (Eq. \ref{eq:28}) is convex
since the Hessian of $\ell\left(\boldsymbol{W}\right)$ is positive
definite (proof in Appendix \ref{subsec:A.3}).

\subsubsection{Generalization to $K$ classes}

We now consider the general multi-label problem with $K>2$ classes.
We consider the maximum number of labels that a training instance
can belongs to, i.e.,

\begin{equation}
\sum_{k}y_{ik}\leq M,\qquad M\leq K.\label{eq:M labels}
\end{equation}

\noindent That means, we observe that a feature vector $\boldsymbol{x}_{i}$
can belong to at most $M$ classes simultaneously in training set.
Therefore, we restrict the label space when training. This restriction
reduces the computational complexity. The generalized multilabel logistic
regression model is

\begin{align}
\Pr\left(\boldsymbol{y}\mid\boldsymbol{x},\boldsymbol{W}\right) & =\frac{1}{Z\left(\boldsymbol{x},\boldsymbol{W}\right)}\prod_{k'=1}^{K}\left\{ \exp\left(y_{k'}\boldsymbol{w}_{k'}^{T}\boldsymbol{x}\right)\right\} ,\label{eq:generalized model}
\end{align}

\noindent where the partition function $Z\left(\boldsymbol{x},\boldsymbol{W}\right)$
is

{\small{}
\begin{equation}
Z\left(\boldsymbol{x},\boldsymbol{W}\right)=\sum_{a_{1}}\exp\left(\boldsymbol{w}_{a_{1}}^{T}\boldsymbol{x}\right)+\sum_{a_{1}}\sum_{a_{2}>a_{1}}\exp\left\{ \left(\boldsymbol{w}_{a_{1}}+\boldsymbol{w}_{a_{2}}\right)^{T}\boldsymbol{x}\right\} +\cdots+\sum_{a_{1}}\cdots\sum_{a_{M}>a_{M-1}}\exp\left\{ \left(\boldsymbol{w}_{a_{1}}+\cdots+\boldsymbol{w}_{a_{M}}\right)^{T}\boldsymbol{x}\right\} .\label{eq:general partition func}
\end{equation}
}{\small\par}

Detailed derivation for the generalized model is presented in Appendix
\ref{subsec:A.1}. The probability of the training set and its negative
log-liklihood are
\begin{equation}
\Pr\left(\left\{ \boldsymbol{y}_{i}\right\} \mid\left\{ \boldsymbol{x}_{i}\right\} ,\boldsymbol{W}\right)=\prod_{i}\left[\frac{1}{Z\left(\boldsymbol{x}_{i},\boldsymbol{W}\right)}\left\{ \prod_{k'=1}^{K}\left(\exp\left(y_{k'}\boldsymbol{w}_{k'}^{T}\boldsymbol{x}\right)\right)\right\} \right],
\end{equation}

\noindent and

\begin{align}
\ell\left(\boldsymbol{W}\right) & =\sum_{i}-\log\Pr\left(\left\{ \boldsymbol{y}_{i}\right\} \mid\left\{ \boldsymbol{x}_{i}\right\} ,\boldsymbol{W}\right)\nonumber \\
 & =\sum_{i}\left[\log Z\left(\boldsymbol{x}_{i},\boldsymbol{W}\right)-\sum_{k'=1}^{K}\left(y_{ik'}\boldsymbol{w}_{k'}^{T}\boldsymbol{x}_{i}\right)\right].\label{eq:general obj func}
\end{align}

It is worth mentioning that regularization terms could also be added
to objective function \ref{eq:general obj func} to prevent overfitting
and a trade off could be found between regularization terms and error
terms.

\subsubsection{Extension to Hilbert Spaces and Kernels}

Logistic regression framework were developed based on the assumption
of linear separability between classes. However, this assumption is
not always the case since there are many problems in which features
are not linearly separable. To address this issue, Kernels were introduced
to map a non-separable dataset into an implicit higher dimensional
reproducing kernel Hilbert space (RKHS) $\mathcal{H}$ known as feature
space where mapped data are linearly separable. Via kernels, it is
possible to compute the inner products of mapped features without
explicitly computing the features in the feature space (Kernel trick)
\cite{Hilbert-Vapnik:1995}. If $\phi:X\rightarrow\mathcal{H}$ is
a mapping from the original space to a feature space, then a kernel
is defined as a function $\kappa:X\times X\rightarrow\mathbb{R}$,
such that for every $x_{i}$,$x_{j}\in X,$

\begin{equation}
\kappa\left(x_{i},x_{j}\right)=\left\langle \phi\left(x_{i}\right),\phi\left(x_{j}\right)\right\rangle ,\label{eq:34}
\end{equation}

\noindent By representer theorem \cite{Representer-KIMELDORF1971},
each weight vector $w_{k}$ can be written as a linear combination
of all projected patterns $\phi\left(x_{i}\right):=\kappa\left(\cdot,x\right)$
in RKHS as

\begin{equation}
w_{k}=\sum_{i=1}^{N}\alpha_{ki}\phi\left(x_{i}\right)\qquad k\in\left\{ 1,\ldots,K\right\} .\label{eq:35}
\end{equation}

\noindent With (Eq. \ref{eq:34}), (Eq. \ref{eq:35}), and the following
notation
\begin{equation}
\boldsymbol{\alpha}_{k}=\left[\alpha_{k1},\alpha_{k2},\cdots,\alpha_{kn}\right]^{T},
\end{equation}

\begin{equation}
K_{i}=\left[\kappa\left(x_{1},x_{i}\right),\kappa\left(x_{2},x_{i}\right),\cdots,\kappa\left(x_{n},x_{i}\right)\right]^{T},
\end{equation}

\noindent The objective function (Eq. \ref{eq:general obj func})
becomes

\begin{align}
\ell\left(\boldsymbol{W}\right) & =\sum_{i}\left[\log Z\left(\boldsymbol{x}_{i},\boldsymbol{W}\right)-\sum_{k'=1}^{K}\left\{ y_{ik'}\boldsymbol{\alpha}_{k'}^{T}K_{i}\right\} \right],
\end{align}

\noindent where

{\small{}
\begin{equation}
Z\left(\boldsymbol{x},\boldsymbol{W}\right)=\sum_{a_{1}=1}^{K}\exp\left(\boldsymbol{\alpha}_{a_{1}}^{T}K_{i}\right)+\sum_{a_{1}=1}^{K}\sum_{a_{2}>a_{1}}^{K}\exp\left\{ \left(\boldsymbol{\alpha}_{a_{1}}+\boldsymbol{\alpha}_{a_{2}}\right)^{T}\boldsymbol{x}\right\} +\cdots+\sum_{a_{1}=1}^{K}\sum_{a_{2}>a_{1}}^{K}\cdots\sum_{a_{M}>a_{M-1}}^{K}\exp\left\{ \left(\boldsymbol{\alpha}_{a_{1}}+\cdots+\boldsymbol{\alpha}_{a_{M}}\right)^{T}\boldsymbol{x}\right\} 
\end{equation}
}{\small\par}

\subsection{Prediction}

In this section, we discuss two different test strategy for the prediction.
The first strategy is the winner-take-all (WTA) method and the second
strategy is the marginal probability method.

The WTA method is conventional decision criterion. In the WTA scheme,
a test pattern is assigned to the class with maximum discriminant
function value. We can apply the WTA method to our MLRH model since
we can compute probabilities for each configuration set of labels
(see, e.g., Eq. \ref{eq:22}, \ref{eq:23}, and \ref{eq:24}). The
weakness of the WTA method is that the maximum number of labels which
can be assigned to a test instance is rectricted by the maximum number
of labels that a traing instance belongs to.

To overcome the weakness of the WTA approach, we employ the marginal
probability method. The marginal probability method enables us to
assign arbitrary number of labels to a test instance by computing
the marginal probability for each label. Therefore, each label can
be assigned independently and probabilistically. Morevover, we can
reduce computation complexity by removing all the comparison procedure
of the WTA method.

\subsubsection{Winner-Take-All}

As discussed previously, the WTA method can also be used for our multilabel
logistic regression model. In this test scheme, a test instance $x_{t}$
is assigned the label configuration $\boldsymbol{y}_{t}$ with maximum
posterior probability $\Pr\left(\boldsymbol{y}_{t}\mid\boldsymbol{x}_{t},\boldsymbol{W}\right)$
The label configurations are restricted to have at most $M$ labels
(Eq. \ref{eq:M labels}). For example, if the number $M=3$ in $K$
classes, we will assign at most $3$ labels to a test instance $\boldsymbol{x}_{t}$
simultaneously. We discuss the WTA method in detail in Appendix \ref{subsec:A.2}.

\subsubsection{Marginal Probability}

\noindent For the second strategy, we use a marginal probability approach
to predict test labels. Assume the number of classes is 2 ($K=2$).
We can compute the marginal probability by using (Eq. \ref{eq:22},
\ref{eq:23}, and \ref{eq:24})

\begin{equation}
\Pr\left(y_{t1}=1\mid\boldsymbol{x}_{t},\boldsymbol{W}\right)=\frac{1}{Z\left(\boldsymbol{x},\boldsymbol{W}\right)}\left[\exp\left(\boldsymbol{w}_{1}^{T}\boldsymbol{x}_{t}\right)+\exp\left\{ \left(\boldsymbol{w}_{1}+\boldsymbol{w}_{2}\right)^{T}\boldsymbol{x}_{t}\right\} \right],
\end{equation}
and

\begin{equation}
\Pr\left(y_{t2}=1\mid\boldsymbol{x}_{t},\boldsymbol{W}\right)=\frac{1}{Z\left(\boldsymbol{x},\boldsymbol{W}\right)}\left[\exp\left(\boldsymbol{w}_{2}^{T}\boldsymbol{x}_{t}\right)+\exp\left\{ \left(\boldsymbol{w}_{1}+\boldsymbol{w}_{2}\right)^{T}\boldsymbol{x}_{t}\right\} \right].
\end{equation}

We will assign each label separately based on the marginal probabilities.
Thresholds for assigning a label would be $0.5$ or they can be also
be obtained by cross validation. We do not put any constraint on the
number of labels that a test instance $\boldsymbol{x}_{t}$ can belong
to simultaneously. Therefore, the benefit of this test scheme is that
we can assign labels in a probabilistic way, even if there are unobserved
label configurations during training.

\section{Majorization-Minimization\label{sec:4}}

In this section, we briefly discuss the Majorization-Minimization
(MM) optimization scheme \cite{Majorization-Hunter2004}. MM is an
optimization framework for both convex and non-convex functions. An
MM procedure operates by iteratively optimizing a surrogate function
that majorizes the objective function. A function $g\left(\theta\mid\theta^{m}\right)$
is said to majorize a real-valued function $f\left(\theta\right)$
at $\theta^{m}$ if

\begin{equation}
g\left(\theta\mid\theta^{m}\right)\geq f\left(\theta\right)\qquad\textrm{for\:all}\:\theta,
\end{equation}

\begin{equation}
g\left(\theta^{m}\mid\theta^{m}\right)=f\left(\theta^{m}\right).
\end{equation}

The surface $\theta\mapsto g\left(\theta\mid\theta^{m}\right)$ lies
above the surface $f(\theta)$ and touches $f\left(\theta\right)$
at point $\theta=\theta^{m}$. Therefore, minimizing or descending
on $g\left(\theta\mid\theta^{m}\right)$ is guaranteed to be a descent
step on $f\left(\theta\right)$. If $\theta^{m+1}$ is the minimizer
of $g\left(\theta\mid\theta^{m}\right)$, then we establish $f\left(\theta^{m+1}\right)=g\left(\theta^{m+1}\mid\theta^{m+1}\right)$
and descend on $g\left(\theta\mid\theta^{m+1}\right)$. This MM algorithm
forces $f\left(\theta\right)$ downhill \cite{Majorization-Hunter2004}.

Hunter and Lange \cite{Majorization-Hunter2004} also presented majorization
via a quadratic upper bound, which can apply to logistic regression.
If a convex function $\kappa\left(\theta\right)$ is twice differentiable
and has bounded curvature, by the mean value theorem, we can majorize
$\kappa\left(\theta\right)$ by

\begin{equation}
\kappa\left(\theta\right)\leq\kappa\left(\theta^{m}\right)+\nabla\kappa\left(\theta^{m}\right)^{T}\left(\theta-\theta^{m}\right)+\frac{1}{2}\left(\theta-\theta^{m}\right)^{T}M\left(\theta-\theta^{m}\right),
\end{equation}

\noindent where $M$ is a positive definite matrix such that $M-\nabla^{2}\kappa\left(\theta\right)$
is nonnegative definite for all $\theta$. Although this quadratic
majorization is not sharp, we can achieve the global minimum since
$\kappa\left(\theta\right)$ is convex. We will show that the quadratic
majorization works reasonably in logistic regression by comparing
to the existing libraries, and then majorize the objective function
(Eq. \ref{eq:general obj func}).

\subsection{MM Algorithm for Multiclass Logistic Regression\label{subsec:4.1}}

In multiclass logistic regression, the objective function is the negative
log-likelihood of the probability of training set (Eq. \ref{eq:multiclass training set prob}),

\begin{equation}
\ell\left(\boldsymbol{W}\right)=\sum_{i=1}^{N}\left[-\sum_{k'=1}^{K}y_{ik'}\boldsymbol{w}_{k'}^{T}\boldsymbol{x}_{i}+\log\sum_{k'=1}^{K}\exp\left(\boldsymbol{w}_{k'}^{T}\boldsymbol{x}_{i}\right)\right].
\end{equation}

\noindent The gradient and Hessian of $\ell\left(W\right)$ with respect
to $w_{k}$ are

\begin{equation}
\boldsymbol{g}\left(\boldsymbol{w}_{k}\right)=\nabla_{\boldsymbol{w}_{k}}\ell\left(\boldsymbol{W}\right)=\sum_{i=1}^{N}\left[y_{ik}+\frac{\exp\left(\boldsymbol{w}_{k}^{T}\boldsymbol{x}_{i}\right)}{\sum_{k'=1}^{K}\exp\left(\boldsymbol{w}_{k'}^{T}\boldsymbol{x}_{i}\right)}\right]\boldsymbol{x}_{i},
\end{equation}

\noindent and

\begin{equation}
\boldsymbol{H}\left(\boldsymbol{w}_{k}\right)=\nabla_{\boldsymbol{w}_{k}}\nabla_{\boldsymbol{w}_{k}^{T}}\ell\left(\boldsymbol{W}\right)=\sum_{i=1}^{N}\left[\frac{\exp\left(\boldsymbol{w}_{k}^{T}\boldsymbol{x}_{i}\right)}{\sum_{k'=1}^{K}\exp\left(\boldsymbol{w}_{k'}^{T}\boldsymbol{x}_{i}\right)}\left(1-\frac{\exp\left(\boldsymbol{w}_{k}^{T}\boldsymbol{x}_{i}\right)}{\sum_{k'=1}^{K}\exp\left(\boldsymbol{w}_{k'}^{T}\boldsymbol{x}_{i}\right)}\right)\right]\boldsymbol{x}_{i}\boldsymbol{x}_{i}^{T}.
\end{equation}

\noindent Let $\boldsymbol{X}$ and $\alpha$ be $\left[\boldsymbol{x}_{1},\boldsymbol{x}_{2},\ldots,\boldsymbol{x}_{N}\right]^{T}$
and $\exp\left(\boldsymbol{w}_{k}^{T}\boldsymbol{x}_{i}\right)/\sum_{k'=1}^{K}\exp\left(\boldsymbol{w}_{k'}^{T}\boldsymbol{x}_{i}\right)$
respectively. Since $0\leq\alpha\left(1-\alpha\right)\leq1/4$, the
Hessian $\boldsymbol{H}\left(\boldsymbol{w}_{k}\right)$ is positive
definite and thus $\frac{1}{4}\boldsymbol{X}^{T}\boldsymbol{X}-\boldsymbol{H}\left(\boldsymbol{w}_{k}\right)$
is nonnegative definite. The corresponding MM algorithm becomes

\begin{equation}
\boldsymbol{w}_{k}^{m+1}\leftarrow\boldsymbol{w}_{k}^{m}-4\left(\boldsymbol{X}^{T}\boldsymbol{X}\right)^{-1}\boldsymbol{g}\left(\boldsymbol{w}_{k}^{m}\right).\label{eq:43}
\end{equation}

\subsubsection{Validity of the MM Algorithm}

We compare the results of the MM algorithm in logistic regression
to those of scikit-learn \cite{library-scikit-learn} and statsmodels
\cite{library-statsmodels} libraries. We use Iris and Wine datasets
for the experiments. Table \ref{tab:5} shows that the MM algorithm
(Eq. \ref{eq:43}) works well comparing to existing libraries.

\noindent 
\begin{table}[H]
\caption{Accuracy on Iris and Wine datasets (\%) in 4-fold cross validation\label{tab:5}}

\centering{}%
\begin{tabular}{c>{\centering}p{2cm}>{\centering}p{2cm}}
\toprule 
\multirow{1}{*}{} & \emph{Iris} & \emph{Wine}\tabularnewline
\midrule
\midrule 
MM algorithm & \textbf{98.02} & 96.08\tabularnewline
\midrule 
Scikit-learn & 97.22 & \textbf{96.11}\tabularnewline
\midrule 
Statsmodels & 97.38 & 95.55\tabularnewline
\bottomrule
\end{tabular}
\end{table}

\subsection{MM Algorithm for Multilabel Logistic Regression}

Derivation of the MM algorithm for the MLRH model is same as described
in Section \ref{subsec:4.1}. The detailed derivation is presented
in Appendix \ref{subsec:A.3}. The MM algorithm for the MLRH model
is

\[
\boldsymbol{w}_{k}^{m+1}\leftarrow\boldsymbol{w}_{k}^{m}-4\left(\boldsymbol{X}^{T}\boldsymbol{X}\right)^{-1}\boldsymbol{g}\left(\boldsymbol{w}_{k}^{m}\right),
\]

\noindent where $\boldsymbol{g}\left(\boldsymbol{w}_{k}^{m}\right)$
is given by

\[
\boldsymbol{g}\left(\boldsymbol{w}_{k}\right)=\nabla_{\boldsymbol{w}_{k}}\ell\left(\boldsymbol{W}\right)=\sum_{i=1}^{N}\left[\frac{\nabla_{\boldsymbol{w}_{k}}Z\left(\boldsymbol{x},\boldsymbol{W}\right)}{Z\left(\boldsymbol{x},\boldsymbol{W}\right)}-y_{ik}\right]\boldsymbol{x}_{i},
\]

\noindent with the partition function $Z\left(\boldsymbol{x},\boldsymbol{W}\right)$
(Eq. \ref{eq:general partition func}).

\section{Experimental Results\label{sec:5}}

This section is devoted to experimental studies that we conducted
to evaluate the performance of our MLRH model. Before presenting and
discussing our experimental results, we briefly discuss some learning
algorithms as baseline methods, and give some information about datasets
and evaluation metrics.

\subsection{Learning Algorithms - Binary Relevance and Label Powerset}

Multilabel classification can be converted into a number of binary
(single-label) problems in a straightforward manner. In \emph{binary
relevance} (BR) learning, a multilabel classification problem is decomposed
into a number of independent binary classification problems. Since
in BR, each binary classifer is trained independently, the correlation
among labels is ignored. Nevertheless, BR usually serves as benchmark
to other multilabel approaches \cite{BR-Luaces2012}. For assigning
a label to a test pattern, it is possible to come up with a probabilistic
scheme for confidence levels and consider a threshold for scores for
which each value or probability above threshold gives rise to a label.
We used BR with Support Vector Machine (SVM) \cite{SVM-Smola2004}
as a benchmark.

In Label Powerset (LP) learning, a multilabel classification problem
with $K$ classes is transformed to a multiclass problem with $2^{K}$
classes. A standard multiclass technique can then be applied to solve
the problem. It is shown that such an approach can give the best empirical
results \cite{LP-Tsoumakas07}. One of the drawbacks of this approach
is that there might be no training patterns corresponding to one or
more combinations of labels and no classifier is learned for them
during training and therefore these label combinations can not be
recognized during testing. We solved transformed LP via multiclass
logistic regression.

\subsection{Dataset}

We tested our MLRH model on two benchmark multilabel datasets, scene
\cite{Scene} and emotions \cite{emotions}. Overview of these datasets
is given in Table \ref{tab:6}.

\noindent 
\begin{table}[H]
\caption{Statistics for Scene and Emotions datasets. Max Labels is the maximum
number of labels that an instance can belong to simultaneously.\label{tab:6}}

\centering{}%
\begin{tabular}{>{\centering}p{1.8cm}>{\centering}p{2.3cm}>{\centering}p{2.3cm}>{\centering}p{2.3cm}>{\centering}p{2.3cm}>{\centering}p{2.3cm}}
\toprule 
Dataset & Domain & \# Instances & \# Attributes & \# Labels & \# Max Labels\tabularnewline
\midrule
\midrule 
Scene & Image & 2407 & 294 & 6 & 3\tabularnewline
\midrule 
Emotions & Music & 593 & 72 & 6 & 3\tabularnewline
\bottomrule
\end{tabular}
\end{table}

\subsection{Evaluation Metrics for Multilabel case}

We use the following evaluation metric:
\begin{itemize}
\item Exact Match Ratio is simply the ratio of correctly classified instances
to all instances. The predicted label of an instance is correct if
it is exactly same as the actual (true) label of the instance.
\[
\textrm{Exact Match Ratio}=\frac{1}{N}\sum_{i=1}^{N}I\left(\tilde{Y_{i}}=Y_{i}\right),
\]
\item Hamming loss \cite{Hamming-Schapire1999} is defined as the fraction
of incorrectly predicted labels to the total number of labels averaged
over all instances and normalized by number of classes,
\[
\textrm{Hamming\:loss}=\frac{1}{NK}\sum_{i=1}^{N}\left|\tilde{Y_{i}}\triangle Y_{i}\right|,
\]
 where $\triangle$ denotes symmetric difference of two sets.
\end{itemize}

\subsection{Results}

We used 5-fold cross validation for training the classifiers. We tuned
the hyperparameters in such a way that BR-SVM performs at its best.
As shown in Table \ref{tab:7}, our MLRH model with WTA test scheme
tends to outperforms other methods in terms of Exact Match Ratio since
our model considers each possibilities of label sets. In marginal
probability test scheme, we simply set the thresholds for each class
to be $0.5$. Therefore, the marginal probability test scheme can
be improved by setting optimal thresholds via cross validation, which
is one of our future works.

\noindent 
\begin{table}[H]
\caption{Experimental results in terms of different evaluation measures via
5-fold cross validation. MLRH stands for Multilabel Logistic Regression
with Hidden Variables model. Winner-take-all (WTA) and marginal probability
(MP) are the test schemes of the MLRH model. RBF denotes RBF kernels.
Exact Match Ratio are in \%.\label{tab:7}}

\centering{}%
\begin{tabular}{c>{\centering}p{2.8cm}>{\centering}p{2.8cm}>{\centering}p{2.8cm}>{\centering}p{2.8cm}}
\toprule 
 & \multicolumn{2}{c}{Scene} & \multicolumn{2}{c}{Emotions}\tabularnewline
\midrule 
Dataset & Exact Match Ratio & Hamming Loss & Exact Match Ratio & Hamming Loss\tabularnewline
\midrule 
MLRH-WTA (RBF) & \textbf{72.61} & 0.079 & \textbf{33.37} & 0.188\tabularnewline
\midrule 
MLRH-MP(RBF) & 54.22 & 0.096 & 31.03 & 0.189\tabularnewline
\midrule 
BR-SVM (RBF) & 64.90 & \textbf{0.075} & 33.20 & \textbf{0.178}\tabularnewline
\midrule 
LP-Multiclass LR & 65.33 & 0.111 & 24.59 & 0.228\tabularnewline
\bottomrule
\end{tabular}
\end{table}

\section{Conclusion and Future Work\label{sec:6}}

In this paper, we proposed a multilabel logistic regression model
via defining a joint distribution of hidden and observed variables.
To our knowledge, it is the first approach to extend the standard
logistic regression by introducing hidden variables. We derived logistic
regression for multiclass from first priciples by using hidden variables
and then obtained multilabel logistic regression using the joint probability
distribution. We also proved that our objective function (negative
log-likelihood) is convex, and thus employ a simple optimization scheme,
majorization-minimization. For the objective function, weight regularization
can be used if needed. The experimental results show that the proposed
multilabel logistic regression model can be competitive comparing
to a set of baseline methods for multilabel learning. For future work,
we will apply our model to different multilabel tasks and compare
other learning algorithms. We will compute the optimal thresholds
of the marginal probability test scheme via cross-validation. We are
considering to deploy the objective function \ref{eq:general obj func}
as a loss in the last layer of a neural network for training multilabel
classifiers.

\bibliographystyle{unsrt}

\appendix

\section{Appendices\label{sec:Appendices}}

\subsection{Generalizing to Multiple Classes\label{subsec:A.1}}

In general case, we have $K$ classes in multi-label setting. Let
$M$ be the maximum number of classes that a feature vector $\boldsymbol{x}$
can belong to simultaneously. We will discuss a simple case, where
$K=3$ and $M=2$, and figure out the general model. We have the joint
probability model

\begin{equation}
\Pr\left(\boldsymbol{y},\boldsymbol{h}\mid\boldsymbol{x},\boldsymbol{W}\right)=\frac{1}{Z\left(\boldsymbol{x},\boldsymbol{W}\right)}\prod_{k}\left\{ \frac{r_{k}+1}{2}\exp\left(h_{k}\boldsymbol{w}_{k}^{T}\boldsymbol{x}\right)\right\} ,
\end{equation}

\noindent where the normalization constant $Z\left(\boldsymbol{x},\boldsymbol{W}\right)$
can computed as

\begin{equation}
Z\left(\boldsymbol{x},\boldsymbol{W}\right)=\sum_{\boldsymbol{y},\boldsymbol{h}}\prod_{k}\left\{ \frac{r_{k}+1}{2}\exp\left(h_{k}\boldsymbol{w}_{k}^{T}\boldsymbol{x}\right)\right\} .
\end{equation}

Let us enumerate all the possible configurations of hidden variables
$\boldsymbol{h}$ and labels $\boldsymbol{y}$ in the joint probability.

\noindent 
\begin{table}[H]

\begin{centering}
\begin{tabular}{ccccccc}
\toprule 
$y_{3}$ & $y_{2}$ & $y_{1}$ & $h_{3}$ & $h_{2}$ & $h_{1}$ & $s\left(\boldsymbol{h},\boldsymbol{x},\boldsymbol{W}\right)$\tabularnewline
\midrule
\midrule 
\multirow{3}{*}{0} & \multirow{3}{*}{0} & \multirow{3}{*}{1} & 0 & 0 & 1 & $\exp\left(\boldsymbol{w}_{1}^{T}\boldsymbol{x}\right)$\tabularnewline
\cmidrule{4-7} \cmidrule{5-7} \cmidrule{6-7} \cmidrule{7-7} 
 &  &  & $\vdots$ & $\vdots$ & $\vdots$ & $\vdots$\tabularnewline
\cmidrule{4-7} \cmidrule{5-7} \cmidrule{6-7} \cmidrule{7-7} 
 &  &  & 1 & 1 & 1 & $0$\tabularnewline
\midrule 
\multirow{3}{*}{0} & \multirow{3}{*}{1} & \multirow{3}{*}{0} & 0 & 1 & 0 & $\exp\left(\boldsymbol{w}_{2}^{T}\boldsymbol{x}\right)$\tabularnewline
\cmidrule{4-7} \cmidrule{5-7} \cmidrule{6-7} \cmidrule{7-7} 
 &  &  & $\vdots$ & $\vdots$ & $\vdots$ & $\vdots$\tabularnewline
\cmidrule{4-7} \cmidrule{5-7} \cmidrule{6-7} \cmidrule{7-7} 
 &  &  & 1 & 1 & 1 & $0$\tabularnewline
\midrule
\multirow{3}{*}{1} & \multirow{3}{*}{0} & \multirow{3}{*}{0} & 1 & 0 & 0 & $\exp\left(\boldsymbol{w}_{3}^{T}\boldsymbol{x}\right)$\tabularnewline
\cmidrule{4-7} \cmidrule{5-7} \cmidrule{6-7} \cmidrule{7-7} 
 &  &  & $\vdots$ & $\vdots$ & $\vdots$ & $\vdots$\tabularnewline
\cmidrule{4-7} \cmidrule{5-7} \cmidrule{6-7} \cmidrule{7-7} 
 &  &  & 1 & 1 & 1 & $0$\tabularnewline
\bottomrule
\end{tabular}\qquad{}%
\begin{tabular}{ccccccc}
\toprule 
$y_{3}$ & $y_{2}$ & $y_{1}$ & $h_{3}$ & $h_{2}$ & $h_{1}$ & $s\left(\boldsymbol{h},\boldsymbol{x},\boldsymbol{W}\right)$\tabularnewline
\midrule
\midrule 
\multirow{3}{*}{0} & \multirow{3}{*}{1} & \multirow{3}{*}{1} & 0 & 1 & 1 & $\exp\left(\boldsymbol{w}_{1}+\boldsymbol{w}_{2}\right)^{T}\boldsymbol{x}$\tabularnewline
\cmidrule{5-7} \cmidrule{6-7} \cmidrule{7-7} 
 &  &  & $\vdots$ & $\vdots$ & $\vdots$ & $\vdots$\tabularnewline
\cmidrule{4-7} \cmidrule{5-7} \cmidrule{6-7} \cmidrule{7-7} 
 &  &  & 1 & 1 & 1 & 0\tabularnewline
\midrule
\multirow{3}{*}{1} & \multirow{3}{*}{0} & \multirow{3}{*}{1} & 1 & 0 & 1 & $\exp\left(\boldsymbol{w}_{1}+\boldsymbol{w}_{3}\right)^{T}\boldsymbol{x}$\tabularnewline
\cmidrule{4-7} \cmidrule{5-7} \cmidrule{6-7} \cmidrule{7-7} 
 &  &  & $\vdots$ & $\vdots$ & $\vdots$ & $\vdots$\tabularnewline
\cmidrule{4-7} \cmidrule{5-7} \cmidrule{6-7} \cmidrule{7-7} 
 &  &  & 1 & 1 & 1 & \tabularnewline
\midrule
\multirow{3}{*}{1} & \multirow{3}{*}{1} & \multirow{3}{*}{0} & 1 & 1 & 0 & $\exp\left(\boldsymbol{w}_{2}+\boldsymbol{w}_{3}\right)^{T}\boldsymbol{x}$\tabularnewline
\cmidrule{4-7} \cmidrule{5-7} \cmidrule{6-7} \cmidrule{7-7} 
 &  &  & $\vdots$ & $\vdots$ & $\vdots$ & $\vdots$\tabularnewline
\cmidrule{4-7} \cmidrule{5-7} \cmidrule{6-7} \cmidrule{7-7} 
 &  &  & 1 & 1 & 1 & $0$\tabularnewline
\bottomrule
\end{tabular}
\par\end{centering}
\caption{Summing over all configuration set of $\boldsymbol{h}$. $s\left(\boldsymbol{h},\boldsymbol{x},\boldsymbol{W}\right)$
is $\prod_{k'=1}^{3}\left\{ \frac{h_{k'}\left(2y_{k'}-1\right)+1}{2}\exp\left(\frac{h_{k'}+1}{2}\boldsymbol{w}_{k'}^{T}\boldsymbol{x}\right)\right\} $.}

\end{table}

\noindent Then, the partition function $Z\left(\boldsymbol{x},\boldsymbol{W}\right)$
becomes

\begin{align}
Z\left(\boldsymbol{x},\boldsymbol{W}\right) & =\exp\left(\boldsymbol{w}_{1}^{T}\boldsymbol{x}\right)+\exp\left(\boldsymbol{w}_{2}^{T}\boldsymbol{x}\right)+\exp\left(\boldsymbol{w}_{3}^{T}\boldsymbol{x}\right)+\exp\left\{ \left(\boldsymbol{w}_{1}+\boldsymbol{w}_{2}\right)^{T}\boldsymbol{x}\right\} \nonumber \\
 & +\exp\left\{ \left(\boldsymbol{w}_{1}+\boldsymbol{w}_{3}\right)^{T}\boldsymbol{x}\right\} +\exp\left\{ \left(\boldsymbol{w}_{2}+\boldsymbol{w}_{3}\right)^{T}\boldsymbol{x}\right\} .
\end{align}

\noindent And some examples of the posterior probabilities $\Pr\left(\boldsymbol{y}\mid\boldsymbol{x},\boldsymbol{W}\right)$
are

\begin{equation}
\Pr\left(y_{1}=1,y_{2}=0,y_{3}=0\mid\boldsymbol{x},\boldsymbol{W}\right)=\frac{1}{Z\left(\boldsymbol{x},\boldsymbol{W}\right)}\exp\left(\boldsymbol{w}_{1}^{T}\boldsymbol{x}\right),
\end{equation}

\begin{equation}
\Pr\left(y_{1}=1,y_{2}=1,y_{3}=0\mid\boldsymbol{x},\boldsymbol{W}\right)=\frac{1}{Z\left(\boldsymbol{x},\boldsymbol{W}\right)}\exp\left\{ \left(\boldsymbol{w}_{1}+\boldsymbol{w}_{2}\right)^{T}\boldsymbol{x}\right\} .
\end{equation}

As we discussed in Section \ref{subsec:3.3}, we can observe that
if $y_{k}=1$, then $\exp\left(\boldsymbol{w}_{k}^{T}\boldsymbol{x}\right)$
term is always included in the posterior probability. With this observation,
we can generalize our Multi-label Logistic Regression with Hidden
variables model

\begin{align}
\Pr\left(\boldsymbol{y}\mid\boldsymbol{x},\boldsymbol{W}\right) & =\frac{1}{Z\left(\boldsymbol{x},\boldsymbol{W}\right)}\prod_{k'=1}^{K}\left\{ \exp\left(y_{k'}\boldsymbol{w}_{k'}^{T}\boldsymbol{x}\right)\right\} ,
\end{align}

\noindent where the partition function $Z\left(\boldsymbol{x},\boldsymbol{W}\right)$
is

{\small{}
\begin{equation}
Z\left(\boldsymbol{x},\boldsymbol{W}\right)=\sum_{a_{1}=1}^{K}\exp\left(\boldsymbol{w}_{a_{1}}^{T}\boldsymbol{x}\right)+\sum_{a_{1}=1}^{K}\sum_{a_{2}>a_{1}}^{K}\exp\left\{ \left(\boldsymbol{w}_{a_{1}}+\boldsymbol{w}_{a_{2}}\right)^{T}\boldsymbol{x}\right\} +\cdots+\sum_{a_{1}=1}^{K}\cdots\sum_{a_{M}>a_{M-1}}^{K}\frac{\exp\left\{ \left(\boldsymbol{w}_{a_{1}}+\cdots+\boldsymbol{w}_{a_{M}}\right)^{T}\boldsymbol{x}\right\} }{M!}.
\end{equation}
}{\small\par}

\subsection{Winner-Take-All\label{subsec:A.2}}

In this test scheme, we restrict the test label space based on the
observed label space in the training set. For a test instance $x_{t}$,
we predict the label with maximum posterior probability. For example,
let us consider the posterior probabilities in two classes

\begin{equation}
\Pr\left(y_{t1}=1,y_{t2}=0\mid\boldsymbol{x}_{t},\boldsymbol{W}\right)=\frac{1}{Z\left(\boldsymbol{x}_{t},\boldsymbol{W}\right)}\exp\left(\boldsymbol{w}_{1}^{T}\boldsymbol{x}_{t}\right),
\end{equation}

\begin{equation}
\Pr\left(y_{t1}=0,y_{t2}=1\mid\boldsymbol{x}_{t},\boldsymbol{W}\right)=\frac{1}{Z\left(\boldsymbol{x}_{t},\boldsymbol{W}\right)}\exp\left(\boldsymbol{w}_{2}^{T}\boldsymbol{x}_{t}\right),
\end{equation}

\begin{equation}
\Pr\left(y_{t1}=1,y_{t2}=1\mid\boldsymbol{x}_{t},\boldsymbol{W}\right)=\frac{1}{Z\left(\boldsymbol{x}_{t},\boldsymbol{W}\right)}\exp\left\{ \left(\boldsymbol{w}_{1}+\boldsymbol{w}_{2}\right)^{T}\boldsymbol{x}_{t}\right\} ,
\end{equation}

\noindent where

\begin{equation}
Z\left(\boldsymbol{x},\boldsymbol{W}\right)=\exp\left(\boldsymbol{w}_{1}^{T}\boldsymbol{x}\right)+\exp\left(\boldsymbol{w}_{2}^{T}\boldsymbol{x}\right)+\exp\left\{ \left(\boldsymbol{w}_{1}+\boldsymbol{w}_{2}\right)^{T}\boldsymbol{x}\right\} .
\end{equation}

\noindent Therefore, we will assign a label set according to the posterior
probabilities. For instance, $y_{t1}=1$ and $y_{t2}=1$ are assigned
if $\exp\left\{ \left(\boldsymbol{w}_{1}+\boldsymbol{w}_{2}\right)^{T}\boldsymbol{x}_{t}\right\} $
is the maximum. Note that in this test scheme, we do not need to consider
the partition function $Z\left(\boldsymbol{x}_{t},\boldsymbol{W}\right)$
since it is the normalization constant of $\Pr\left(\boldsymbol{y}_{t}\mid\boldsymbol{x}_{t},\boldsymbol{W}\right)$.
This test scheme can be simplified. Assume the case that $y_{t1}=1$
and $y_{t2}=1$ are assigned. That means

\begin{align}
\exp\left\{ \left(\boldsymbol{w}_{1}+\boldsymbol{w}_{2}\right)^{T}\boldsymbol{x}_{i}\right\}  & \geq\exp\left(\boldsymbol{w}_{1}^{T}\boldsymbol{x}_{i}\right)
\end{align}

\begin{align}
 & \Rightarrow\exp\left(\boldsymbol{w}_{1}^{T}\boldsymbol{x}_{i}\right)\exp\left(\boldsymbol{w}_{2}^{T}\boldsymbol{x}_{i}\right)\geq\exp\left(\boldsymbol{w}_{1}^{T}\boldsymbol{x}_{i}\right)\nonumber \\
 & \Rightarrow\exp\left(\boldsymbol{w}_{2}^{T}\boldsymbol{x}_{i}\right)\geq1\\
 & \Rightarrow\boldsymbol{w}_{2}^{T}\boldsymbol{x}_{i}\geq0\nonumber 
\end{align}

\noindent In this way, we can simplify the WTA method. For example,
$y_{t1}=1$ and $y_{t2}=1$ are assigned if $\boldsymbol{w}_{1}^{T}\boldsymbol{x}_{i}\geq0$
and $\boldsymbol{w}_{2}^{T}\boldsymbol{x}_{i}\geq0$.

\subsection{MM Algorithm for MLHR model\label{subsec:A.3}}

From the objective function (Eq. \ref{eq:general obj func}), the
gradient of $\ell\left(W\right)$ with respect to $w_{k}$ are

\begin{equation}
\boldsymbol{g}\left(\boldsymbol{w}_{k}\right)=\nabla_{\boldsymbol{w}_{k}}\ell\left(\boldsymbol{W}\right)=\sum_{i=1}^{N}\left[\frac{\nabla_{\boldsymbol{w}_{k}}Z\left(\boldsymbol{x},\boldsymbol{W}\right)}{Z\left(\boldsymbol{x},\boldsymbol{W}\right)}-y_{ik}\right]\boldsymbol{x}_{i}.
\end{equation}

\noindent Let $\alpha$ be equal to $\nabla_{\boldsymbol{w}_{k}}Z\left(\boldsymbol{x},W\right)/Z\left(\boldsymbol{x},W\right)$.
Then Hessian is

\begin{align}
\boldsymbol{H}\left(\boldsymbol{w}_{k}\right) & =\nabla_{\boldsymbol{w}_{k}}\nabla_{\boldsymbol{w}_{k}^{T}}\ell\left(\boldsymbol{W}\right)=\sum_{i=1}^{N}\left[\alpha\left(1-\alpha\right)\right]\boldsymbol{x}_{i}\boldsymbol{x}_{i}^{T}\nonumber \\
 & =\boldsymbol{X}^{T}\boldsymbol{S}\boldsymbol{X}
\end{align}

\noindent where $\boldsymbol{X}=\left[\boldsymbol{x}_{1},\boldsymbol{x}_{2},\ldots\boldsymbol{x}_{N}\right]^{T}$,
$S=\textrm{diag\ensuremath{\left[\alpha\left(1-\alpha\right)\right]}}$,
and $Z(x,W)$ is given by (Eq. \ref{eq:general partition func}).
The Hessain $\boldsymbol{H}\left(\boldsymbol{w}_{k}\right)$ is positive
definite since

\noindent 
\begin{equation}
\boldsymbol{u}^{T}\boldsymbol{H}\left(\boldsymbol{w}_{k}\right)\boldsymbol{u}=\boldsymbol{u}^{T}\boldsymbol{X}^{T}\boldsymbol{S}\boldsymbol{X}\boldsymbol{u}=\left(\boldsymbol{X}\boldsymbol{u}\right)^{T}\boldsymbol{S}\left(\boldsymbol{X}\boldsymbol{u}\right)>0,\qquad\forall\alpha\neq0.
\end{equation}
Therefore, our objective function (Eq. \ref{eq:general obj func})
is convex. Furthermore, $\frac{1}{4}\boldsymbol{X}^{T}\boldsymbol{X}-\boldsymbol{H}\left(\boldsymbol{w}_{k}\right)$
is nonnegative definite since $\boldsymbol{H}\left(\boldsymbol{w}_{k}\right)\leq\frac{1}{4}\boldsymbol{X}^{T}\boldsymbol{X}$.
The corresponding MM algorithm becomes

\begin{equation}
\boldsymbol{w}_{k}^{m+1}\leftarrow\boldsymbol{w}_{k}^{m}-4\left(\boldsymbol{X}^{T}\boldsymbol{X}\right)^{-1}\boldsymbol{g}\left(\boldsymbol{w}_{k}^{m}\right).
\end{equation}

\end{document}